\documentclass[letterpaper]{article} 
\usepackage{aaai2026}  
\usepackage{times}  
\usepackage{helvet}
\usepackage{courier}  
\usepackage[hyphens]{url}  
\usepackage{graphicx} 
\usepackage{natbib} 
\usepackage{caption}
\usepackage{algorithm}
\usepackage{algorithmic}

\usepackage{newfloat}
\usepackage{listings}
\DeclareCaptionStyle{ruled}{labelfont=normalfont,labelsep=colon,strut=off} 
\newfloat{listing}{tb}{lst}{}
\usepackage{graphicx}
\usepackage{amsmath}
\usepackage{amssymb}
\usepackage{booktabs}
\usepackage{tabularx}
\usepackage{bm}
\usepackage{float}
\usepackage{xcolor}
\usepackage[
  colorlinks=true,
  linkcolor=black, 
  citecolor=black,
  urlcolor=blue
]{hyperref}
\usepackage{cleveref}
\Crefname{figure}{Fig.}{Figs.}
\crefname{figure}{Fig.}{Figs.}
\usepackage{calc}
\usepackage[svgnames,x11names,table]{xcolor}
\usepackage{multirow}
\usepackage{tikz}
\usepackage{mdframed}
\usepackage{makecell}
\usetikzlibrary{positioning}

\definecolor{iccvblue}{rgb}{0.21,0.49,0.74}
\definecolor{myred}{HTML}{B85450}
\definecolor{myblue}{HTML}{6C8EBF}
\definecolor{polaris}{RGB}{121,150,196}
\definecolor{composite}{RGB}{230,163,126}
\definecolor{flickr8k-cf}{RGB}{129,190,142}
\definecolor{flickr8k-ex}{RGB}{211,123,126}
\definecolor{TitleColor}{gray}{0.95}
\definecolor{LightPink}{HTML}{FAE6E7}

\newcommand{\change}[1]{\textcolor{black}{#1}}

\DeclareMathAlphabet{\mymathbb}{U}{BOONDOX-ds}{m}{n}

\title{LLM-Free Image Captioning Evaluation in Reference-Flexible Settings}
\author{
    Shinnosuke Hirano,
    Yuiga Wada, 
    Kazuki Matsuda,
    Seitaro Otsuki,
    Komei Sugiura
}
\affiliations{
    Keio University\\
    \{shinhirano, yuiga, k2matsuda0, otsu8sei14, komei.sugiura\}@keio.jp  
}

\begin{document}

\maketitle
\begin{abstract}
We focus on the automatic evaluation of image captions in both reference-based and reference-free settings. Existing metrics based on large language models (LLMs) favor their own generations; therefore, the neutrality is in question. Most LLM-free metrics do not suffer from such an issue, whereas they do not always demonstrate high performance. To address these issues, we propose Pearl, an LLM-free supervised metric for image captioning, which is applicable to both reference-based and reference-free settings. We introduce a novel mechanism that learns the representations of image--caption and caption--caption similarities. 
Furthermore, we construct a human-annotated dataset for image captioning metrics, that comprises approximately 333k human judgments collected from 2,360 annotators across over 75k images. 
Pearl outperformed other existing LLM-free metrics on the Composite, Flickr8K-Expert, Flickr8K-CF, Nebula, and FOIL datasets in both reference-based and reference-free settings.
Our project page is available at \url{https://pearl.kinsta.page/}.
\end{abstract}

\section{Introduction}
\begin{figure}[t]
    \centering
    \includegraphics[width=\linewidth]{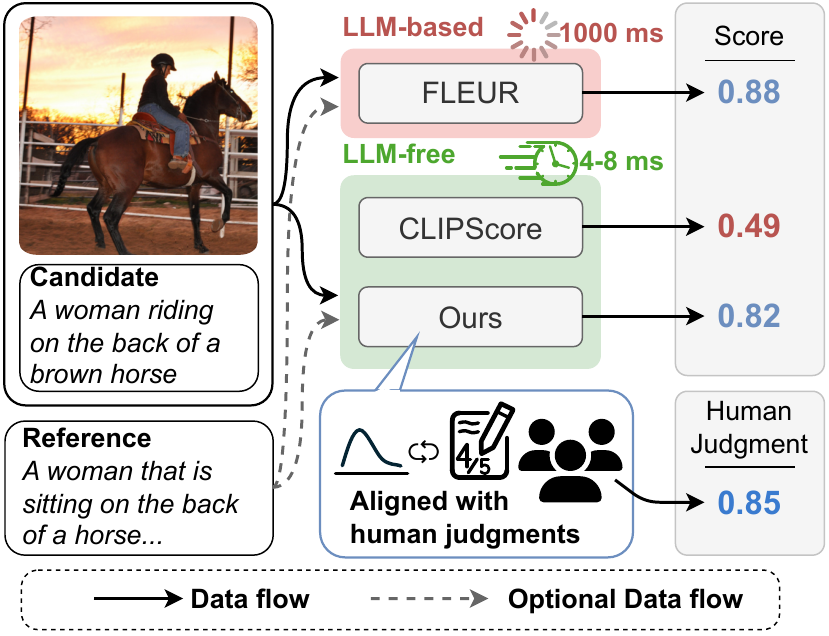}
    \caption{Pearl is an LLM-free automatic evaluation metric for image captioning. Pearl works significantly faster than slow LLM-based metrics; thus it is suitable for fast development cycles of practical image captioning models. Moreover, Pearl is the first LLM-free supervised metric that can handle both reference-based and reference-free evaluation with a single supervised model.
}
    \label{fig:eye-catch}
    
\end{figure}
Image captioning has been widely studied and utilized in various applications, including assisting visually impaired people and providing explanations in robotics. 
The efficient development of image captioning models relies on automatic evaluation metrics that closely align with human judgments. Recent efforts have focused on building metrics that correlate more strongly with human evaluations than classic metrics such as CIDEr~\cite{cider}.
Although this task has been widely investigated, it remains challenging even for state-of-the-art (SOTA) metrics (e.g.~\citet{yao2024hifi, expert}).
Indeed, the correlations of these metrics with human judgments are still lower than the correlations among human judgments \cite{lee2024fleur, flickr}.
Therefore, constructing aligned metrics would be beneficial for both the image captioning community and society.

Previous studies have shown that existing metrics based on LLMs favor their own generations~\cite{llmfavor, aibias}.
Moreover, they often suffer from prohibitively long inference times, which makes their application impractical.
Therefore, it is important to develop an LLM-free metric that does not suffer from the neutrality and speed issues.

LLM-free metrics \cite{pac-s, deneb, sarto2024positive} are significantly faster than LLM-based metrics. However, they exhibit several notable issues.
For example, although supporting both reference-based and reference-free evaluation is important for practical applications \cite{tong2024gveval}, few demonstrate high performance in both settings.
This is partially because metrics that support both settings \cite{clipscore, sarto2024positive} simply compute cosine similarity between embeddings from encoders such as CLIP~\cite{clip}, which can be less aligned with human judgments~\cite{polos}.
Moreover, supervised metrics employ fixed representations for similarity (e.g. RUSE \cite{ruse}), which may be suboptimal because of the lack of representation learning.

To address these issues, we propose Pearl, a supervised automatic evaluation metric for image captioning.
Fig.~\ref{fig:eye-catch} shows an overview of Pearl.
In addition to its LLM-free nature, Pearl differs from existing approaches in several respects.
First, unlike other learning-based metrics \cite{umic, polos}, Pearl is a single model capable of evaluating captions in both reference-based and reference-free settings.
Second, we introduce the Adaptive RUSE-type Similarity Mechanism to learn the representations of image--caption and caption--caption similarities.

To train Pearl, we construct a new dataset, Spica, which includes 333K human judgments for a diverse collection of image--caption pairs. 
The Spica dataset stands out with its significantly larger number of human judgments and greater diversity of both images and annotators compared with other standard datasets \cite{composite, flickr, umic, polos, deneb}.
Specifically, it includes approximately 333K human judgments, which is 2.5 times the number of human judgments in the dataset previously recognized as having the largest number of human judgments.

Our key contributions are as follows:
\begin{itemize}
\item We introduce the Image-Guided Evaluation Module and the Reference-Guided Evaluation Module to handle the similarities, which enables both reference-based and reference-free evaluation with a single model.
\item We constructed Spica, a publicly available dataset for image captioning metrics, which is significantly larger than previous datasets. Specifically, it contains 75,535 images and 333,397 human judgments from 2,360 annotators.
\item We achieved SOTA performance on standard benchmarks within LLM-free metrics, enabling practical large-scale candidate caption evaluation.
\end{itemize}

\begin{figure*}[t]
    \centering
    \includegraphics[width=\linewidth]{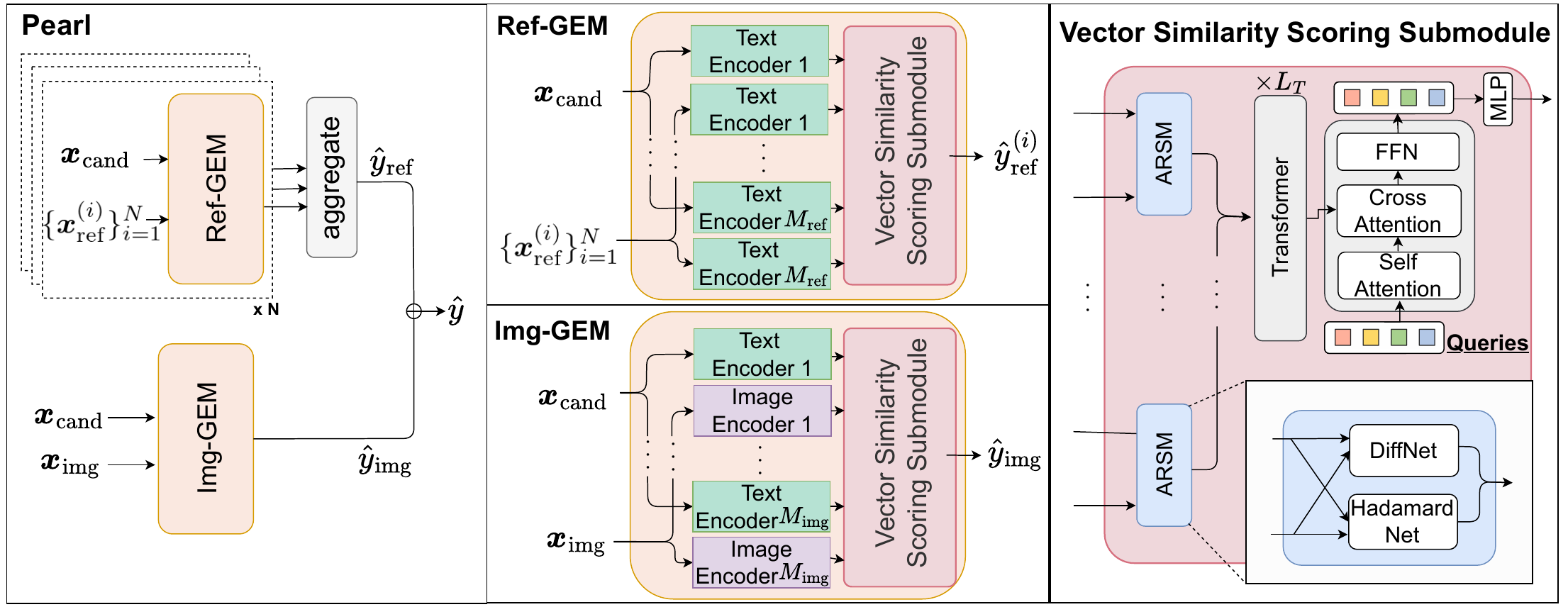}
    
    \caption{Overview of the proposed metric, Pearl.  Our proposed metric consists of the Img-GEM and multiple Ref-GEMs. The Img-GEM computes a score for a candidate caption based on the associated image, whereas each Ref-GEM calculates a score based on a reference. In reference-based setting, Pearl computes the scores for candidate caption based on either the image or the references, and then fuses them into the final prediction score. Conversely, in the reference-free setting, the final prediction score is computed based solely on the image.
}
    \label{fig:model}
    
\end{figure*}

\section{Related Work}
\subsection{LLM-free Metrics}
\paragraph{Reference-based metrics.}
Standard reference-based metrics include BLEU~\cite{bleu}, METEOR~\cite{meteor}, ROUGE~\cite{rouge}, CIDEr~\cite{cider}, and SPICE~\cite{spice, jaspice}, which evaluate captions using rule-based approaches based only on reference captions.  

In contrast, several reference-based metrics also incorporate images.
Notable metrics include ViLBERTScore~\cite{vilbertscore}, which feeds image--caption and caption--caption pairs into ViLBERT~\cite{vilbert} and calculates the cosine similarities. 
Recent metrics~\cite{polos, deneb} adopt a supervised approach, directly learning to evaluate captions from human judgments.
These metrics are based on the RUSE-type similarity~\cite{ruse, comet}, which merely computes fixed representations of the element-wise difference and Hadamard product.

\paragraph{Reference-free metrics.}
Reference-free metrics such as CLIPScore~\cite{clipscore} rely only on images for evaluating candidate captions.
CLIPScore calculates scores by computing the cosine similarity between the embeddings of CLIP~\cite{clip}.
Sarto et al.~\cite{pac-s} highlighted that CLIP was trained on alt-texts, whose text distribution significantly differs from that of image captions; they therefore proposed PAC-S and PAC-S++~\cite{sarto2024positive}, variants of CLIPScore, which employ CLIP finetuned by generated images and captions.
BLIP2Score~\cite{zeng2024meacap} evaluates candidate captions in a manner similar to that of CLIPScore by using embeddings of BLIP-2~\cite{blip2}.

\subsection{LLM-based Metrics} 
LLM-based metrics use LLM or MLLM in its evaluation process.
Various studies~\cite{chan2023clair, lee2024fleur, ohi2024harmoniceval, tong2024gveval, yao2024hifi, Zeng_2024hicescore, expert, vela}  have proposed LLM/MLLM-based metrics for image captioning.
CLAIR~\cite{chan2023clair} is the first metric that leverages LLMs in a zero-shot manner to compare references and candidate captions.
While CLAIR evaluates the captions without considering images, FLEUR~\cite{lee2024fleur} utilizes LLaVA~\cite{llava, llava-1.5} to incorporate images. FLEUR introduces score smoothing, which calibrates the raw score by the probabilities associated with output tokens.
Similarly, EXPERT~\cite{expert} first produces the evaluation score using score smoothing, and then generates a brief explanation based on three criteria --- \textit{fluency}, \textit{relevance}, and \textit{descriptiveness}. On the other hand, HiFi-Score~\cite{yao2024hifi} converts captions and images into hierarchical parsing graphs by LLMs, and compares them through fine-grained graph matching.

\section{Methodology}
We propose Pearl, an LLM-free supervised automatic evaluation metric for image captioning, which is applicable to both reference-based and reference-free settings. 
Pearl adopts a single-model strategy that jointly addresses both settings, facilitating the learning of shared representations.
In contrast to approaches that train separate models for each setting, this design allows all samples -- whether reference-based or reference-free -- to contribute to training.

The inputs to Pearl are an image $\bm{x}_{\mathrm{img}}$, a candidate caption $\bm{x}_{\mathrm{cand}}$, and a set of references $\bm{X}_{\mathrm{ref}}$. In the reference-based setting,
$\bm{X}_{\mathrm{ref}} = \{\bm{x}_{\mathrm{ref}}^{(i)}\}_{i=1}^N$,
where $N \ge 1$ and each reference $\bm{x}_{\mathrm{ref}}^{(i)}$ is represented as a binary matrix
$\bm{x}_{\mathrm{ref}}^{(i)} \in \{0, 1\}^{V \times L}$.
In the reference-free setting, $\bm{X}_{\mathrm{ref}} = \varnothing$, corresponding to $N = 0$.
Here, $V$ and $L$ represent the vocabulary size and the number of tokens in a caption, respectively.
In both settings, $\bm{x}_{\mathrm{img}} \in \mathbb{R}^{3 \times H \times W}$ represents the input image, where $H$ and $W$ denote the height and width of the image, respectively. The candidate caption $\bm{x}_{\mathrm{cand}} \in \{0, 1\}^{V \times L}$ is also encoded as a binary matrix.
Given these inputs, Pearl outputs the final evaluation score $\hat{y}$ in an LLM-free manner, with or without reference captions.

\paragraph{Why do we need LLM-free metrics?} LLM-based metrics have two severe limitations. 
First, recent studies have shown that LLM-based evaluators favor their own generations~\cite{llmfavor, aibias}; therefore, their neutrality is in question.
Second, as will be discussed in the \textit{Experiments} section, LLM-based metrics often suffer from prohibitively long inference times, making their application impractical.
Therefore, it is important to develop an LLM-free metric that does not suffer from the neutrality and speed issues.

\subsection{Architecture Overview}

Pearl consists of four main features: Adaptive RUSE-type Similarity Mechanism (ARSM), Image-Guided Evaluation module (Img-GEM), Reference-Guided Evaluation module (Ref-GEM), and Vector Similarity Scoring (VSS) submodule.
\cref{fig:model} shows an overview of Pearl.

The core components of our proposed method are ARSM and VSS.
ARSM can handle learnable representations of image--caption and caption--caption similarities, unlike previous metrics which rely on fixed representations \cite{comet, ruse, polos, deneb}.
ARSM can be applicable to existing supervised metrics for natural language generation~\cite{ruse, comet}.
In contrast, VSS generates an intermediate score for candidates from multiple perspectives by fusing embeddings from multiple encoders (e.g., CLIP, BLIP-2, BEiT-3).
Specifically, it computes ARSM-based representations for each encoder independently and then fuses them to produce an intermediate score.

Building on these cores, the proposed metric consists of Img-GEM and $N$ Ref-GEMs, each of which internally invokes VSS, where $N$ denotes the number of references.
First, Img-GEM evaluates candidate captions based on images using VSS, while each Ref-GEM assesses them based on reference captions, also using VSS.
Subsequently, Pearl generates the final score by combining the $N$ Ref-GEM outputs with the Img-GEM outputs.
Unlike previous supervised metrics~\cite{polos, deneb} that perform early fusion of features, we adopt a late-fusion approach, allowing effective learning in both reference-based and reference-free settings.

\subsection{Adaptive RUSE-type Similarity Mechanism}

We introduce Adaptive RUSE-type Similarity Mechanism (ARSM), which learns representations of image--caption and caption--caption similarities.
Previous approaches (e.g. \citet{ruse}) often have utilized element wise differences and the Hadamard product as fixed representations for the similarity.
This RUSE-type similarity operation has demonstrated its effectiveness in various tasks, including machine translation \cite{ruse, comet} and image captioning \cite{polos, deneb}.
However, such fixed operations may be suboptimal because they do not involve representation learning.

Although replacing the Hadamard product and element wise difference with feed-forward networks sounds theoretically plausible, this does not always yield effective results (see ablation studies in the \textit{Experiments} section). 
Therefore, we construct feed-forward networks that satisfy the same input and output requirements as the Hadamard product and element wise difference.  We then integrate this network in a trainable form into our proposed metric. We introduce two components of ARSM: DiffNet and HadamardNet.

\paragraph{DiffNet.}
DiffNet consists of a single layer of feed-forward networks. 
DiffNet replaces the difference operation between two features, $\bm{x}_1 \in \mathbb{R}^{n\times d}$ and $\bm{x}_2 \in \mathbb{R}^{n\times d}$. The difference operation is easily captured by a single fully-connected layer with the following initialization parameters:
$\bm{o}_{\mathrm{diff}} = \bm{W}_{1}\bm{x}_1 + \bm{W}_{2}\bm{x}_2 + \bm{b},$
where $\bm{o}_{\mathrm{diff}}$ denotes the output of DiffNet, $\mathbf{W}_{1}$ and $\mathbf{W}_{2}$ are weight parameters, and $\bm{b}$ is a bias parameter.
We initialize $\mathbf{W}_1 = \mathbf{1}_{d \times n}$, $\mathbf{W}_2 = - \mathbf{1}_{d \times n}$, and $\bm{b} = \mathbf{0}$, where $\mathbf{1}_{d \times n}$ denotes $d \times n$ matrix of ones.

\paragraph{HadamardNet.}
HadamardNet, composed of $L_h$ CNN layers, is used to capture the relationship between $\bm{x}_1 $ and $\bm{x}_2 $ instead of the Hadamard product operations. 
Given the challenges in constructing mappings of the Hadamard products using methods akin to DiffNet, we employ a learning-based approach.
Initially, we generate multiple vectors from a uniform distribution $U(a,b)$ to collect training samples. We then pretrain HadamardNet to learn the Hadamard products of these vectors. Here, $a$ and $b$ denotes the hyperparameters.
We initialize HadamardNet with the parameters obtained from this pretraining. 
The final output of HadamardNet is $\bm{o}_{\mathrm{had},L_h}$.
Finally, the output of ARSM is defined as $\bm{o}_{\mathrm{ARSM}} = [\bm{o}_{\mathrm{diff}, L_d}; \; \bm{o}_{\mathrm{had}, L_h}]$.

\subsection{Img-GEM and Ref-GEM}
Img-GEM and Ref-GEM handle image--caption and caption--caption similarities, respectively.
Previous studies \cite{clipscore, pac-s, lee2024fleur} have noted that preparing high-quality references is extremely time-consuming and have proposed metrics for both settings \cite{lee2024fleur, pac-s}. 
Although supervised metrics outperform other LLM-free metrics on standard benchmarks, almost all existing supervised metrics \cite{polos, deneb} assume a reference-based setting.
This limitation arises because these metrics fuse the features of a candidate caption, references, and an image early in the process.
In contrast, our supervised metric employs a late-fusion approach, computing scores from Img-GEM and multiple Ref-GEMs and then fusing these into the final prediction score.

Img-GEM computes a score for a candidate caption based on the associated image, while each Ref-GEM calculates a score based on a reference.
Initially, in Img-GEM, we extract image features
$\{\, \bm{v}_{\mathrm{vgem}, j} \in \mathbb{R}^{d_{\mathrm{vgem}, j}} \,|\, j=1, 2, \dots, M_{\mathrm{img}}\}$
from $\bm{x}_{\mathrm{img}}$ using $M_{\mathrm{img}}$ image encoders. Here, $d_{\mathrm{vgem}, j}$ denotes the dimension of the $j$-th encoder in Img-GEM. Subsequently, we also extract sentence embeddings $\{\, \bm{c}_{\mathrm{vgem}, j} \in \mathbb{R}^{d_{\mathrm{vgem}, j}} \,|\, j=1, 2, \dots, M_{\mathrm{img}}\}$ from $\bm{x}_{\mathrm{cand}}$ using the corresponding $M_{\mathrm{img}}$ text encoders.
We use \change{frozen} CLIP \cite{clip}, BLIP-2\cite{blip2}, and BEiT-3 \cite{beit3} as the image and text encoders in Img-GEM because they are known for their ability to align image and text features.

The $i$-th Ref-GEM processes the $i$-th reference $\bm{x}_{\mathrm{ref}}^{(i)}$ and extracts sentence embeddings
$\{\, \bm{r}^{(i)}_{\mathrm{rgem}, j} \in \mathbb{R}^{d_{\mathrm{rgem}, j}} \,|\, j=1, 2, \dots, M_{\mathrm{ref}}\}$
using $M_{\mathrm{ref}}$ text encoders. Here, $d_{\mathrm{rgem}, j}$ denotes the dimension of the $j$-th encoder in Ref-GEM.
Similarly, we extract sentence embeddings
$\{\, \bm{c}_{\mathrm{rgem}, j} \in \mathbb{R}^{d_{\mathrm{rgem}, j}} \,|\, j=1, 2, \dots, M_{\mathrm{ref}}\}$
from $\bm{x}_{\mathrm{cand}}$ using $M_{\mathrm{ref}}$ text encoders. As the text encoders in Ref-GEM, we use \change{frozen} BLIP-2, BEiT-3, and Stella \cite{zhang2024jasper}, a lightweight sentence embedding model with approximately 400M parameters.
Here, Stella is employed because its performance is comparable to that of SOTA LLMs on standard benchmarks (e.g. \citet{zhang2024jasper}).

Subsequently, in Img-GEM, we feed the extracted
$\{\, (\bm{c}_{\mathrm{vgem}, j}, \bm{v}_{\mathrm{vgem}, j}) \,|\, j=1, 2, \dots, M_{\mathrm{img}}\}$
into the VSS submodule. Moreover, Ref-GEM feeds the extracted
$\{\, (\bm{c}_{\mathrm{rgem}, j}, \bm{r}^{(i)}_{\mathrm{rgem}, j}) \,|\, j=1, 2, \dots, M_{\mathrm{img}} \}$
into the same submodule. 
Finally, the Img-GEM output and the $i$-th Ref-GEM output are $\bm{h}_{\mathrm{img}}$ and $\bm{h}_{\mathrm{ref}}^{(i)}$, respectively.

\subsection{Vector Similarity Scoring Submodule}
VSS submodule evaluates $\bm{x}_{\mathrm{cand}}$ based on the inputs $\{ (\bm{c}_i, \bm{h}_i) \mid i = 1, \dots, M \}$, 
where $\bm{c}_i$ and $\bm{h}_i$ denote the $i$-th embedding of $\bm{x}_{\mathrm{cand}}$, and the $i$-th embedding used as a basis for evaluating $\bm{x}_{\mathrm{cand}}$, respectively. Here, $M$ denotes the number of encoders.
By fusing embeddings obtained from multiple encoders, VSS can assess $\bm{x}_{\mathrm{cand}}$ from multiple perspectives.
Initially, this submodule employs ARSM on each pair $(\bm{c}_i, \bm{h}_i)$ to obtain features $\bm{g}_i$ that capture representations of differences between embeddings.

These features $\{\bm{g}_i\}_{i=1}^{M}$ are then concatenated and fed into a Transformer consisting of $L_T$ layers to extract features $\bm{g}_{\mathrm{enc}}$ that are beneficial for evaluation.
Inspired by \cite{blip2}, we then feed $\bm{g}_{\mathrm{enc}}$ into a Q-Former, which contains $L_Q$ layers, to obtain the feature $\bm{g}_{\mathrm{dec}}$.
Finally, the score $\hat{y}_{vc}$ is obtained by processing $\bm{g}_{\mathrm{dec}}$ through an MLP.

In the reference-based setting, we compute the scores for $\bm{x}_{\mathrm{cand}}$ based on either $\bm{x}_{\mathrm{img}}$ or $\{\bm{x}_{\mathrm{ref}}^{(i)}\}_{i=1}^N$ and then fuse them into the final prediction score. In the reference-free setting, the final prediction score for $\bm{x}_{\mathrm{cand}}$ is computed based solely on the image.
The image-based score for $\hat{y}_{\mathrm{img}}$ and the reference-based score $\hat{y}_{\mathrm{ref}}$ are computed as follows:
\begin{align}
\hat{y}_{\mathrm{img}} &= \sigma\bigl(\mathrm{MLP}(\bm{h}_{\mathrm{img}})\bigr), \\
\hat{y}_{\mathrm{ref}} &= \max_{i}\Bigl(\sigma\bigl(\mathrm{MLP}(\bm{h}^{(i)}_{\mathrm{ref}})\bigr)\Bigr),
\end{align}
where $\sigma$ denotes the sigmoid function.

Finally, the final evaluation score $\hat{y}$ is computed by fusing  $\hat{y}_{\mathrm{img}}$ and $\hat{y}_{\mathrm{ref}}$ as follows:
\begin{align}
\hat{y} =
\begin{cases}
\lambda \hat{y}_{\mathrm{img}} + (1 - \lambda) \hat{y}_{\mathrm{ref}} & \text{(Reference-based)} \\
\hat{y}_{\mathrm{img}} & \text{(Reference-free)}
\end{cases},
\end{align}
where $\lambda$ is a hyperparameter within $[0, 1]$.
We set $\lambda$ to $0.5$.
We use the Huber loss $\mathcal{L}_{\mathrm{huber}}(\cdot, \cdot)$ for its robustness to outliers.
In the reference-based setting, the loss is computed as the average of $\mathcal{L}_{\mathrm{huber}}(y, \hat{y}_{\mathrm{img}})$ and $\mathcal{L}_{\mathrm{huber}}(y, \hat{y}_{\mathrm{ref}})$.
In the reference-free setting, we use only $\mathcal{L}_{\mathrm{huber}}(y, \hat{y}_{\mathrm{img}})$.

\begin{table*}[t]
\centering
\small 
\setlength{\tabcolsep}{1mm} 
\begin{tabular}{l c cc  cc  cc  cc  cc  c} \toprule
\multirow{2}{*}{\textbf{Metrics}} & \multirow{2}{*}{\textbf{Ref}} & \multicolumn{2}{c}{\textbf{Composite}}& \multicolumn{2}{c}{\textbf{Flickr8K-Ex}} & \multicolumn{2}{c}{\textbf{Flickr8K-CF}} & \multicolumn{2}{c}{\textbf{Nebula}} & \multicolumn{2}{c}{\textbf{FOIL}} & \textbf{Test time} \\ 
                         &                      & $\tau_b$ & $\tau_c$ & $\tau_b$ & $\tau_c$ & $\tau_b$ & $\tau_c$ & $\tau_b$ & $\tau_c$ & {\footnotesize 1-ref [\%]} & {\footnotesize 4-ref [\%]} & [hour]\\ \midrule

\rowcolor{gray!10}
\multicolumn{13}{l}{\textbf{LLM-free methods (reference-based)}} \\

BLEU \cite{bleu}        & \checkmark & 28.3  & 30.6  & 30.6  & 30.8  & 16.4  & 8.7   & 46.5  & 44.1  & 66.5  & 82.6  & $<$ 0.01\\
ROUGE \cite{rouge}      & \checkmark & 30.0  & 32.4  & 32.1  & 32.3  & 19.9  & 10.3  & 45.8  & 43.4  & 71.7  & 79.3  & $<$ 0.01 \\
CIDEr \cite{cider}      & \checkmark & 34.9  & 37.7  & 43.6  & 43.9  & 24.6  & 12.7  & 51.5  & 48.8  & 82.5  & 90.6  & $<$ 0.01 \\
METEOR \cite{meteor}    & \checkmark & 36.0  & 38.9  & 41.5  & 41.8  & 22.2  & 11.5  & 50.2  & 47.6  & 78.8  & 82.6  & $<$ 0.01 \\
SPICE \cite{spice}      & \checkmark & 38.8  & 40.3  & 51.7  & 44.9  & 24.4  & 12.0  & 51.5  & 47.4  & 75.5  & 86.1  & 0.089 \\
RefCLIP-S $^\dagger$ \cite{clipscore} & \checkmark & 51.2 & 55.4 & 52.6	& 53.0 & 36.4 & 18.8 & 53.6 & 50.8 & 91.0 & 92.6 & 0.014  \\
RefPAC-S $^\dagger$ \cite{pac-s} & \checkmark & 53.0  & 57.3  & 55.5  & 55.9  & 37.6  & 19.5  & 54.7  & 51.9  & 93.7  & 94.9  & 0.023 \\
Polos \cite{polos}      & \checkmark & 53.7  & 57.6  & 56.1  & 56.4  & 37.8  & 19.5  & 58.0  & 55.0  & 93.3  & 95.4  & 0.036 \\
Ref-HICEScore \cite{Zeng_2024hicescore} & \checkmark & 53.9  & 58.7  & \underline{57.2}  & \underline{57.7}  & \underline{38.2}  & \underline{19.8}  & -     & -     & \underline{96.4}  & \underline{97.0}  & - \\
\textsc{Deneb} $^\dagger$ \cite{deneb} & \checkmark & 54.0 & 57.9 & 55.6 & 56.5 & 38.0 & 19.6 & \underline{58.1} & \underline{55.1} & 95.1 & 96.1 & 0.038 \\
RefPAC-S++ $^\dagger$ \cite{sarto2024positive} & \checkmark & \underline{54.7} & \underline{59.1} & 55.3 & 55.7 & 37.9 & 19.6 & 53.3 & 50.6 & 93.5 & 94.1 & 0.023\\
\textbf{Ours (ViT-B/32)}        & \checkmark & \textbf{55.8}  & \textbf{60.4}  & \textbf{58.2}  & \textbf{58.6}  & \textbf{38.6}  & \textbf{20.0}  & \textbf{58.4}  & \textbf{55.4}  & \textbf{96.5}  & \textbf{97.2}  & 0.043\\ \midrule

\rowcolor{gray!10}
\multicolumn{13}{l}{\textbf{LLM-free methods (reference-free)}} \\
CLIP-S \cite{clipscore}   &   & 49.8  & 53.8  & 51.1  & 51.2  & 34.4  & 17.7  & 50.5  & 47.9  & 87.2  & 87.2  & $<$ 0.01 \\
PAC-S $^\dagger$ \cite{pac-s} &   & 51.5 & 55.7  & 53.9  & 54.3  & 36.0  & 18.6  & 51.0  & 48.3  & 89.9  & 89.9  & \underline{0.013} \\
HICEScore \cite{Zeng_2024hicescore} &   & 53.1  & 57.9  & \underline{55.9}  & \underline{56.4}  & \underline{37.2}  & \underline{19.2}  & -     & -     & 93.1  & 93.1  & -\\
PAC-S++ $^\dagger$ \cite{sarto2024positive} &   & 53.9  & 58.3  & 54.1  & 54.5  & 37.0  & 19.1  & 50.5  & 47.9  & 90.2  & 90.2  & \underline{0.013} \\
BLIP2Score \cite{zeng2024meacap} &   & \textbf{56.9}  & \textbf{61.5}  & 52.2  & 52.5  & 36.7  & 19.0  & \underline{53.0}  & \underline{50.7}  & \underline{94.3}  & \underline{94.3}  & 0.020 \\
\textbf{Ours (ViT-B/32)}                  &   & \underline{54.0}  & \underline{58.4}  & \textbf{56.2}  & \textbf{56.6}  & \textbf{37.8}  & \textbf{19.5}  & \textbf{55.9}  & \textbf{53.0}  & \textbf{96.7}  & \textbf{96.7}  & 0.043 \\ \midrule

\rowcolor{gray!10}
\multicolumn{13}{l}{\textbf{LLM-based methods}} \\

CLAIR\footnotemark[1] \cite{chan2023clair}
& \checkmark & -     & 61.0  & 58.3 &  48.8  & 38.2  & 17.0    & -     & -     & -  & 93.6 & 8.3  \\
FLEUR \cite{lee2024fleur} &   & -     & 63.5  & -     & 53.0  & 38.6  & -     & -     & -     & 96.8  & 96.8  & 3.7  \\
Ref-FLEUR \cite{lee2024fleur} & \checkmark & -     & 64.2  & -     & 51.9  & 38.8  & -     & -     & -     & 97.3  & 98.4  & 4.0  \\
HiFiScore \cite{yao2024hifi} &   & -     & 65.7  & -     & 58.4  & -     & -     & -     & -     & -     & -     & -\\ 
Ref-HiFiScore \cite{yao2024hifi} & \checkmark & -     & 65.8  & -     & 58.4  & -     & -     & -     & -     & -     & -     & - \\ 
G-VEval \cite{tong2024gveval} &   & - & - & 61.5 & 59.7 & 38.7 & 20.2 & - & - & - & - & 11 \\
Ref-G-VEval \cite{tong2024gveval} & \checkmark & - & - & 60.5 & 58.7 & 38.2 & 19.9 & - & - & 97.8 & 98.4 & 11\\
EXPERT \cite{expert}&   & - & 65.0 & - & 56.7 & 39.3 & - & - & 54.9 & - & - & -\\
\bottomrule
\end{tabular}
\caption{Quantitative comparison with baseline metrics.
The column ``Ref'' indicates whether the method uses reference captions.
Bold font indicates the best values and underline indicates the second best values. ``-'' indicates either non executable code or unavailable data. 
``$\tau_b$'' and ``$\tau_c$'' represent Kendall's $\tau_b$ and $\tau_c$ correlation coefficients, respectively. ``Test time'' refers to the total inference time for evaluating the test sets of COCO~\cite{coco}, nocaps~\cite{nocaps}, and TextCaps~\cite{textcaps}.
Metrics marked with $^\dagger$ use ViT-B/32 as the backbone for a fair comparison.}

\label{tab:quantitative}
\end{table*}

\section{Spica Dataset}
The development of supervised metrics for image captioning requires large-scale and diverse datasets in terms of the number of images, annotators, and captioning models. 
Specifically, a dataset should: (i) contain an extensive collection of human judgments and (ii) include a diverse and large-scale set of images to effectively capture the similarities between images and candidate captions.
However, few standard datasets~\cite{composite, flickr, polos, deneb} meet both criteria.
The dataset with the largest collection of human judgments~\cite{polos} comprises approximately 131k human judgments but contains only 13k images. 
As noted in~\cite{deneb}, this imbalance could potentially lead to suboptimal evaluation of various types of images.
Similarly, Nebula~\cite{deneb} has the largest number of unique images but only 32k human judgments, which can result in suboptimal alignment with human judgments.

To address these gaps, we constructed the Spica dataset, which includes (i) a large number of human judgments and (ii) a diverse and extensive collection of images. 
The Spica dataset comprises 333,397 human judgments collected from 2,360 annotators and 75,535 unique images.
Our dataset stands out with its significantly larger number of human judgments and greater diversity of both images and annotators.
Specifically, our dataset includes 2.5 times the number of human judgments as the Polaris dataset~\cite{polos}, the largest existing dataset in terms of human judgments.
Moreover, our dataset contains 2.3 times as many unique images as the Nebula dataset~\cite{deneb}, which currently has the highest number of unique images.
To generate candidate captions, we employed the ten image captioning models.
\change{The criteria of model selection, the statistical information and details of our dataset can be found in the Appendix.}

\begin{figure*}[t]
    \centering
    \includegraphics[width=\linewidth]{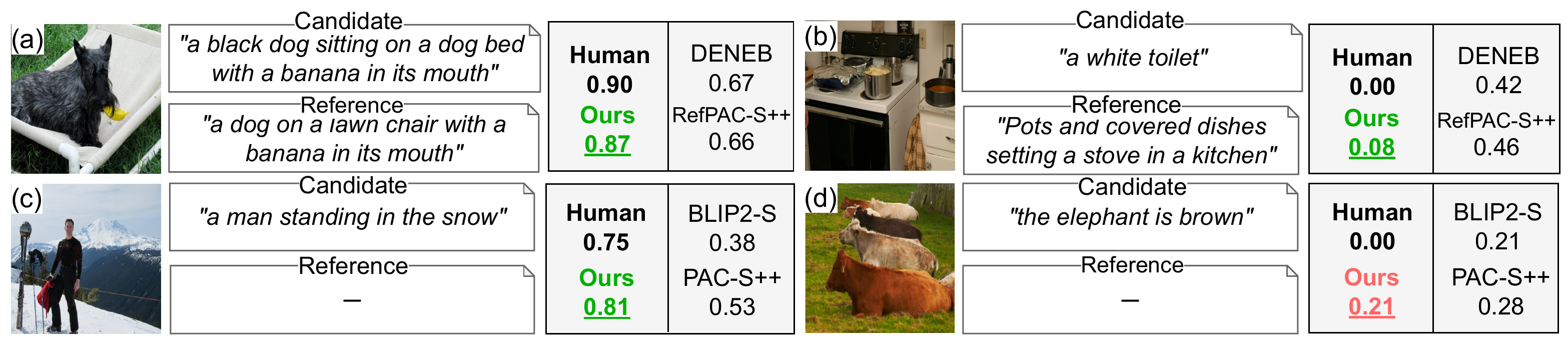}
    
    \caption{Qualitative results on the Nebula dataset. Cases (a) and (b) show successful cases in the reference-based setting, while Case (c) highlights successful sample in the reference-free setting. Case (d) shows a failure case in the reference-free setting.
    }
    \label{fig:qualitive}  
\end{figure*}

\section{Experiments}
\label{experimental_setup}
To evaluate the correlation between the metric and human judgments \change{on both in-domain and out-of- domain samples}, we used the Composite~\cite{composite}, Flickr8K-Expert~\cite{flickr}, Flickr8K-CF, and Nebula~\cite{deneb} datasets.
Note that we corrected the annotation errors found in the Nebula dataset.
Moreover, to assess robustness against hallucination, we conducted comparative experiments with the FOIL dataset~\cite{foil}. 
We selected these benchmarks because they are standard in this field.
Details of the baselines are provided in the Appendix.

\footnotetext[1]{Following standard practice~\cite{tong2024gveval, lee2024fleur}, we report the scores of CLAIR as reproduced in~\cite{tong2024gveval, lee2024fleur}, because the originally reported results exhibit a disparity in the calculation method compared to other metrics (\url{https://github.com/DavidMChan/clair/issues/2}).}

\subsection{Quantitative Results}

\paragraph{Human correlation.}
\Cref{tab:quantitative} presents the quantitative comparison results between the baselines and our proposed metric on the Composite, Flickr8K-Expert, Flickr8K-CF, and Nebula datasets. 
We employed Kendall’s $\tau_b$ and $\tau_c$ correlation coefficients, as they are standard for evaluation in this field.
In the reference-based setting,  Pearl achieved $\tau_b$ scores of 55.8, 58.2, 38.6, and 58.4 and $\tau_c$ scores of 60.4, 58.6, 20.0 and 55.4 on the Composite, Flickr8K-Expert, Flickr8K-CF, and Nebula datasets, respectively. 
These results indicate that Pearl outperformed the existing LLM-free baseline metrics by margins of 1.1, 1.0, 0.4 and 0.3 points for $\tau_b$ and 1.3, 0.9, 0.2, and 0.3 points for $\tau_c$.
Pearl achieved an average increase of 1.1 points in $\tau_b$ and 0.93 in $\tau_c$ over Ref-HICEScore, the previous SOTA baseline metrics.
In comparison, the average improvement of Ref-HICEScore over RefPAC-S++, the second best previous metrics, was only 0.47 and 0.60 in $\tau_b$ and $\tau_c$, respectively. These results indicate that Pearl yielded substantial improvements.

In the reference-free setting, our method achieved $\tau_b$ of 56.2, 37.8, and 55.9, and $\tau_c$ of 56.6, 19.5, and 53.0 on the Flickr8K-CF, Flickr8K-Expert, and Nebula datasets, respectively. 
These results indicate improvements of 0.3, 0.6, and 2.9 points for $\tau_b$, and 0.2, 0.3, and 2.3 points for $\tau_c$ over the previous LLM-free SOTA baseline metrics, respectively.

\paragraph{Sensitivity to hallucination.}
\Cref{tab:quantitative} shows the performance of the proposed metric and the baseline metrics on the FOIL dataset.
In the reference-based setting, Pearl achieved SOTA performance among LLM-free metrics, with scores of 96.5\% in the one-reference setting and 97.2\% in the four-references setting. 
Moreover, in the reference-free setting, our metric also achieved a SOTA performance of 96.7\% among LLM-free metrics, representing an improvement of 2.4 points over existing LLM-free metrics. 
These results suggest that our metric is robust even against hallucinations.

\paragraph{Inference time.}
\label{sec:inference_time}
The rightmost column of \Cref{tab:quantitative} shows the total inference time for evaluating the test sets of COCO~\cite{coco}, nocaps~\cite{nocaps}, and TextCaps\cite{textcaps}, which are the standard benchmarks for image captioning~\cite{mm1, florence2}. 
All measurements were taken on a system equipped with a GeForce RTX 3090 and an Intel Core i9-10900KF.
The inference times for recent LLM-free metrics, RefCLIP-S and RefPAC-S, were 0.014 and 0.023 hours, respectively. 
Pearl showed an inference time of 0.043 hours, suggesting that Pearl is comparable in terms of inference speed. 
In contrast, LLM-based metrics such as CLAIR and FLEUR exhibited significantly longer inference times of 8.3 and 3.7 hours, respectively, which makes them at least 85 times slower than Pearl.
\begin{table*}[t]
\centering
\setlength{\tabcolsep}{1mm} 
\begin{tabular}{c c c c c c c c c c}
\toprule
  \multirow{2}{*}{\textbf{Metric}} 
 & \multirow{2}{*}{\textbf{ARSM}}
 & \multicolumn{3}{c}{\textbf{Image Encoder}} 
 & \textbf{Text Encoder} 
 & \textbf{Composite}
 & \textbf{Flickr8K-Ex}
 & \textbf{Flickr8K-CF}
 & \textbf{Nebula}\\
\cmidrule(lr){3-5} \cmidrule(lr){7-7} \cmidrule(lr){8-8} \cmidrule(lr){9-9} \cmidrule(lr){10-10} \cmidrule(lr){6-6}
   &  & CLIP & BLIP-2 & BEiT-3 & Stella
 & $\tau_c$ & $\tau_c$ & $\tau_b$ & $\tau_c$ \\
\midrule

  (i)   & none        & \checkmark & \checkmark & \checkmark & \checkmark & 30.9 & 31.2 & 17.4 & 41.7 \\
  (ii)  & initial     & \checkmark & \checkmark & \checkmark & \checkmark & 54.3 & 58.2 & 38.5 & 52.5 \\
  (iii) & RUSE-type   & \checkmark & \checkmark & \checkmark & \checkmark & 59.5 & 56.7 & 37.3 & 50.3 \\
 (iv)  & adaptive    & \checkmark & \checkmark & \checkmark &            & 51.9 & 56.0 & 37.3 & 52.0 \\
  (v)   & adaptive    &            & \checkmark & \checkmark & \checkmark & 55.5 & 54.2 & 35.8 & 53.9 \\
  (vi)  & adaptive    & \checkmark &            & \checkmark & \checkmark & 59.0 & 57.3 & 38.0 & 54.0 \\
  (vii) & adaptive    & \checkmark & \checkmark &            & \checkmark & 59.9 & 58.3 & 38.4 & 53.9 \\
  (viii)& adaptive    & \checkmark & \checkmark & \checkmark & \checkmark & \textbf{60.4} & \textbf{58.6} & \textbf{38.6} & \textbf{55.4} \\
\bottomrule
\end{tabular}
\caption{
Results of ARSM ablation and feature extractor ablation in reference-based setting. ``$\tau_b$'' and ``$\tau_c$'' represents the Kendall’s $\tau_b$ and $\tau_c$ correlation coefficient, respectively. Results for reference-free setting are in the Appendix.}
\label{tab:combined_ablation}
\end{table*}
\subsection{Qualitative Results}
\paragraph{Successful cases.}
\cref{fig:qualitive} shows successful cases of Pearl on the Nebula dataset.
In the figure, Cases (a) and (b) show successful cases in the reference-based setting, while Case (c) displays a successful case in the reference-free setting.

In Case (a), 
the human judgment was 0.90 because $\bm{x}_{\mathrm{cand}}$ appropriately describes the image.  
Ref-CLIPS and Ref-PACS++ incorrectly evaluated this sample with scores of 0.67 and 0.66, respectively, whereas Pearl appropriately evaluated it with a value of 0.87. Similarly, in Case (b), Pearl yielded the score most aligned with human judgment.

Case (c) shows a successful sample in the reference-free setting. The human judgment was $0.75$ because the caption only partially describes the image.
Both PAC-S++ and BLIP2-S incorrectly evaluated the sample with scores of 0.53 and 0.38, respectively, whereas Pearl assigned a score of 0.81, which more closely aligned with the human judgment.

\paragraph{Failure case.}
Case (d) in \cref{fig:qualitive} illustrates a failure case in the reference-free setting. In this sample, $\bm{x}_{\mathrm{cand}}$ was incorrectly described as ``The elephant is brown,'' which does not correspond to $\bm{x}_{\mathrm{img}}$. This discrepancy resulted in a human judgment score of $0.00$. 
However, our proposed metric evaluated this sample with a score of $0.21$.
Similarly, PAC-S++ and BLIP2-S assigned the sample scores of $0.28$ and $0.21$, respectively, demonstrating discrepancies as well. Given that the subject (i.e., ``elephant'') is misidentified in this example, the presence of the adjective ``brown'' should not contribute to an increased score. 
However, these metrics include color information when computing similarity; therefore, we hypothesize that the failure occurs because even if the core part (e.g., the subject identification) is incorrect, the presence of correct color information leads to a score increase.

\change{Moreover, to examine additional failure cases, we analyzed the 100 samples with the greatest absolute differences between $\hat{y}$ and $y$. For details of the analysis, see Appendix.}
\subsection{Ablation Study}
\label{ablation_study}
We conducted ablation studies to assess the impact of each module and Spica dataset.
Tables \ref{tab:combined_ablation} and \ref{tab:dataset_ablation} show the results of the ablation studies.
Results in the reference-free setting are provided in the Appendix.

\paragraph{ARSM ablation.}
We investigated the contribution of ARSM, which learns features, in three different ways: by removing it, freezing its parameters at their initial values, and replacing it with a non learnable RUSE-type similarity.
\Cref{tab:combined_ablation} shows that Metric (i) yielded correlation coefficients of 30.9, 31.2, 17.4, and 41.7 for Composite, Flickr8K-Expert, Flickr8K-CF, and Nebula, respectively. 
These values decreased by 29.5, 27.4, 21.2, and 13.7 points compared with those of Metric (viii). 

Moreover, Metric (iii), a version where ARSM was replaced with a non-learnable RUSE-type similarity, yielded correlation coefficients 
of 59.5, 56.7, 37.3, and 50.3 for Composite, Flickr8K-Expert, Flickr8K-CF, and Nebula, respectively. 
These values decreased by 0.9, 1.9, 1.3, and 5.1 points compared to Metric (viii). 
These results indicate that ARSM was effective at extracting significant features for evaluation and significantly contributed to performance. 
\begin{table}[t]
\centering
\begin{tabular}{llccc}
\toprule
 \multirow{2}{*}{\textbf{Metrics}} &  \multirow{2}{*}[\dimexpr -0.5ex]{\shortstack[t]{\textbf{Training}\\\textbf{Dataset}}}  & \textbf{Com} & \textbf{Ex} & \textbf{CF}  \\  
\cmidrule(lr){3-3} \cmidrule(lr){4-4} \cmidrule(lr){5-5}
& & $\tau_c$ & $\tau_c$ & $\tau_b$  \\ \midrule
(a) DENEB & Nebula  & 57.9  & 56.5 & 38.0  \\
(b) DENEB & Spica  & 58.6  & 56.9 & 38.2  \\
(c) Pearl & Nebula  & 59.2  & 57.7 & 37.9  \\
(d) Pearl & Spica  & \textbf{60.4}  & \textbf{58.6} & \textbf{38.6}  \\
\midrule
\end{tabular}
\caption{Results of the Spica ablation. ``Com'', ``EX'' and ``CF'' represent Composite, Flickr8K-EX and Flickr8K-CF. }
\label{tab:dataset_ablation}
\end{table}
\paragraph{Feature extractor ablation.}
We investigated the contribution of image and text encoders by removing CLIP, BLIP-2, BEiT-3, and Stella.
\Cref{tab:combined_ablation} demonstrates that Metric (iv), which corresponds to Pearl without the Stella encoder, achieved correlation coefficients of 51.9, 56.0, 37.3, and 52.0 on Composite, Flickr8K-Expert, Flickr8K-CF, and Nebula, respectively.
Compared to the original Pearl (viii), these scores decreased by 8.5, 2.6, 1.3, and 3.4 points, respectively, highlighting the significant contribution of Stella to performance in the reference-based setting.

Moreover, \Cref{tab:combined_ablation} shows that Metric (v), which corresponds to Pearl without the CLIP encoder, yielded correlation coefficients of 55.5, 54.2, 35.8 and 53.9 on Composite, Flickr8K-Expert, Flickr8K-CF, and Nebula, respectively.
Compared with those of the original Pearl (viii), these values decreased by 4.9, 4.4, 2.8 and 1.5 points.
These results indicate that CLIP significantly contributed to the performance.

However, they do not indicate that CLIP is the sole contributor to Pearl’s performance.
To further investigate this, we conducted an ablation study where all encoders except CLIP were removed.
Details are provided in the Appendix.

\paragraph{Spica Ablation.}
We evaluated the effectiveness of the Spica dataset by training the existing supervised metric on Spica.
We compared DENEB \cite{deneb} because it was the previous SOTA among supervised LLM-free metrics.
As shown in Table~\ref{tab:dataset_ablation}, DENEB achieved better performance when trained on Spica~(b) compared to when trained on Nebula~(a), which is their original training dataset.
These results indicate the versatility of Spica for supervised metrics.
Moreover, comparing Pearl trained on Nebula (c) with the original model (d) highlights the contribution of the Spica dataset to overall performance.
These indicate that Spica provided effective data for training supervised metrics.

\section{Conclusion}
We focused on the automatic evaluation for image captioning.
Our contributions are as follows: 
(i) we proposed Pearl, a supervised metric for reference-flexible settings,
(ii) we introduced Img-GEM and Ref-GEM to handle image--caption and caption--caption similarities,
(iii) we introduced ARSM, which learns rich similarity representations,
(iv) we constructed Spica, a dataset with diverse images and extensive human judgments, 
and (v) we achieved SOTA performance on standard benchmarks within LLM-free metrics.

The proposed metric performed well on standard benchmarks, although limitations remain.
Our error analysis revealed that Pearl assigned incorrect scores to captions that lacked descriptiveness.
Details of the analysis are provided in the Appendix.

\section{Acknowledgments}
This work was supported by a grant from Apple, Inc. Any views, opinions, findings, and conclusions or recommendations expressed in this material are those of the authors and should not be interpreted as reflecting the views, policies, or position, either expressed or implied, of Apple, Inc.
This work was also partially supported by JSPS KAKENHI Grant Number 23K28168 and JST Moonshot.

\bibliography{aaai2026}
\clearpage
\section{Appendix A: Additional Related Works}
\paragraph{Benchmarks.} Several benchmarks have been proposed for image captioning metrics~\cite{flickr, composite, polos, deneb, foil, an2025can}.
Flickr8K-Expert~\cite{flickr} and Flickr8K-CF~\cite{flickr} provide human judgments for candidate captions associated with 1,000 images from the Flickr8K test set. Flickr8K-Expert and Flickr8K-CF contain 5,822 and 47,830 candidate captions, respectively. 
In Composite~\cite{composite}, each image is paired with two candidate captions and one reference. Composite has 3,996 unique images and 11,985 human judgments for each candidate caption.
The Polaris dataset~\cite{polos} contains over 131k human judgments for the captions generated for approximately 14k images, making it the largest among similar datasets in terms of human judgments. Similarly, Nebula~\cite{deneb} is the largest in terms of image count, containing approximately three times as many images as Polaris.
In this study, we constructed Spica, a significantly larger human-annotated dataset for image captioning metrics than previous ones.
Our dataset contains 2.5 times the number of human judgments that Polaris has and  2.3 times the number of images that Nebula has (see the \textit{Spica Dataset} section).

\section{Appendix B: Datasets}
\subsection{Spica Dataset}
\label{appendix:spica}

\paragraph{Image captioning models.}
We employ the following standard image captioning models to generate candidate captions: VinVL~\cite{VinVL}, GRIT~\cite{GRIT}, $\mathcal{M}^2$-Transformer~\cite{m2trm}, BLIP~\cite{BLIP}, GIT~\cite{GIT}, OFA~\cite{OFA}, and BLIP-2~\cite{blip2}. For BLIP, we utilized two versions that employ ViT-B and ViT-L~\cite{ViT} as their image encoders. Similarly, for BLIP-2, we employed two variants that use Flan-T5~\cite{flant5} and OPT~\cite{OPT} as the LLMs.
\change{We selected these models in terms of their providing organizations (e.g., Alibaba, Microsoft, Academia) and release years, in order to minimize the model bias.}

\paragraph{Image diversity.}
\change{To analyze the diversity of images in our dataset, we measure semantic coverage based on the geometric coverage in the CLIP image embedding space.
Specifically, we computed the proportion of the convex hull volume formed by each dataset (Flickr8k-CF \cite{flickr}, Flickr8k-EX, Composite \cite{composite}, and Nebula \cite{deneb}) to the entire OpenImages.
The results show that Flickr8k-CF, Flickr8k-Expert, Composite, Nebula, and Spica cover 25.3\%, 25.3\%, 50.3\%, 85.4\%, and 91.6\% of the OpenImages hull.
These results demonstrate that Spica provided more diverse images than other standard benchmarks \cite{flickr,composite,deneb}.
}

\paragraph{Annotation bias.}
\change{
Spica was designed to exhibit low bias in the distribution of human judgments, comparable to those of Composite and Polaris.
By contrast, in Flickr8k-CF and Flickr8k-EX, 88\% and 55\% of the scores are concentrated at 0.0, respectively.
Moreover, Krippendorff’s $\alpha$ for the inter-annotator agreement was 0.44, which is higher than that of standard datasets (e.g. CapEval1k~\cite{umic}).
}

\paragraph{Annotations.}
We followed standard practices in this field during the annotation process~\cite{composite, thumb, umic, polos, deneb}.
Human judgments were collected on a five-point scale to evaluate the appropriateness of candidates in relation to the given images and references. 
Annotators were instructed to assess captions across three dimensions: \textit{fluency}, \textit{relevance}, and \textit{descriptiveness}. 
To ensure data reliability, annotations from evaluators displaying suspicious behavior, such as abnormally short response times or consistently uniform ratings, were excluded.
We normalized the human judgments from a five-point scale to the range $[0,1]$.

\paragraph{Statistics.}
The statistics of the Spica dataset are summarized in Table~\ref{tab:spica}.
The total number of references in Spica is 1,128,933, with a vocabulary size of 36,947, a total word count of 11,278,014, and an average sentence length of 10.0 words. 
The total number of candidates is 333,397, with a vocabulary size of 15,697, a total word count of 2,936,768, and an average sentence length of 8.8 words. 
We divided the Spica dataset into training, validation, and test sets, containing 296,149, 3,000, and 3,287 samples, respectively.
We used the training set, validation set, and test set to train the model, tune the hyperparameters, and evaluate the model's performance, respectively.

\begin{table}[]
\centering
\begin{tabular}{lrrr}
\toprule
Dataset           & Annotators & {\begin{tabular}{c}Unique \\ images\end{tabular}} & {\begin{tabular}{c}Human \\ judgments\end{tabular}} \\ \midrule
CapEval1K   & 5          & 250           & 1,000           \\
THumB        & 4          & 500           & 2,500           \\
MMHE       & 5          & 100           & 4,500           \\
Composite    & --          & 3,996         & 11,985          \\
flickr8k-EX & 21         & 1,000         & 16,992          \\
Nebula     & 805        & 32,978        & 32,978          \\
flickr8K-CF  & --          & 1,000         & 47,830          \\
Polaris   & 550        & 13,691        & 131,020         \\
$\textbf{Spica (Ours)}$             & $\textbf{2,360}$      & $\textbf{75,535}$        & $\textbf{333,397}$        \\ \bottomrule
\end{tabular}
\caption{Comparison between Spica and the standard datasets. Spica stands out with its significantly larger number of human judgments and greater diversity of both images and annotators.}
\label{tab:spica}
\end{table}

\begin{table*}[t]
\centering
\small 
\setlength{\tabcolsep}{1mm} 
\begin{tabular}{l c cc  cc  cc  cc  cc  c} \toprule
\multirow{2}{*}{\textbf{Metrics}} & \multirow{2}{*}{\textbf{Ref}} & \multicolumn{2}{c}{\textbf{Composite}}& \multicolumn{2}{c}{\textbf{Flickr8K-Ex}} & \multicolumn{2}{c}{\textbf{Flickr8K-CF}} & \multicolumn{2}{c}{\textbf{Nebula}} & \multicolumn{2}{c}{\textbf{FOIL}} & \textbf{Test time} \\ 
                         &                      & $\tau_b$ & $\tau_c$ & $\tau_b$ & $\tau_c$ & $\tau_b$ & $\tau_c$ & $\tau_b$ & $\tau_c$ & {\footnotesize 1-ref [\%]} & {\footnotesize 4-ref [\%]} & [hour]\\ \midrule

\rowcolor{gray!10}
\multicolumn{13}{l}{\textbf{LLM-free (reference-based)}} \\

BERTScore \cite{bertscore} & \checkmark & 30.2 & 30.1  & 37.8  & 46.7  & 22.8  & 11.5  & 47.5  & 47.1  & 88.6  & 92.1  & 0.040 \\
BARTScore \cite{bartscore} & \checkmark & 30.4  & 43.5  & 32.5  & 37.8  & 24.3  & 10.3  & 47.4  & 45.0  & 85.3  & 91.1  & 0.68 \\
SPARCS \cite{smurf}     & \checkmark & 41.4  & 43.1  & 43.8  & 48.1  & 25.2  & 13.0  & 53.9  & 50.0  & 86.0  & 92.1  & $<$ 0.01 \\
\textbf{Ours (ViT-B/32)}        & \checkmark & \textbf{55.8}  & \textbf{60.4}  & \textbf{58.2}  & \textbf{58.6}  & \textbf{38.6}  & \textbf{20.0}  & \textbf{58.4}  & \textbf{55.4}  & \textbf{96.5}  & \textbf{97.2}  & 0.043\\ \midrule

\rowcolor{gray!10}
\multicolumn{13}{l}{\textbf{LLM-free (reference-free)}} \\
BRIDGE   \cite{bridge}             &   & 52.9  & 57.2  & 55.4  & 55.8  & 36.3  & 19.0  & -     & -     & 93.0  & 93.0  & -\\
\textbf{Ours (ViT-B/32)}                  &   & \textbf{54.0}  & \textbf{58.4}  & \textbf{56.2}  & \textbf{56.6}  & \textbf{37.8}  & \textbf{19.5}  & \textbf{55.9}  & \textbf{53.0}  & \textbf{96.7}  & \textbf{96.7}  & 0.043 \\

\bottomrule
\end{tabular}
\caption{Additional quantitative comparison with baselines.
The column ``Ref'' indicates whether the method uses reference captions.
Bold font indicates the best values. ``-'' indicates either non executable code or unavailable data. 
``$\tau_b$'' and ``$\tau_c$'' represent Kendall's $\tau_b$ and $\tau_c$ correlation coefficients, respectively. ``Test time'' refers to the total inference time for evaluating the test sets of COCO~\cite{coco}, nocaps~\cite{nocaps}, and TextCaps~\cite{textcaps}.
}
\label{tab:additional_baseline}
\end{table*}

\begin{table*}[t]
\centering
\small
\begin{tabular}{c c c c c c c c c c}
\toprule
  \multirow{2}{*}{\textbf{Metric}} 
 & \multirow{2}{*}{\textbf{ARSM}}
 & \multicolumn{3}{c}{\textbf{Image Encoder}} 
 & \textbf{Text Encoder} 
 & \textbf{Composite}
 & \textbf{Flickr8K-Ex}
 & \textbf{Flickr8K-CF}
 & \textbf{Nebula}\\
\cmidrule(lr){3-5} \cmidrule(lr){6-6} \cmidrule(lr){7-7} \cmidrule(lr){8-8} \cmidrule(lr){9-9} \cmidrule(lr){10-10} 
 &  & CLIP & BLIP-2 & BEiT-3 & Stella
 & $\tau_c$ & $\tau_c$ & $\tau_b$ & $\tau_c$ \\
\midrule

  (i)   & none       & \checkmark & \checkmark & \checkmark & \checkmark & 30.7 & 30.4 & 16.7 & 42.3 \\
  (ii)  & initial    & \checkmark & \checkmark & \checkmark & \checkmark & 53.0 & 56.2 & 37.6 & 52.2 \\
 (iii) & RUSE-type  & \checkmark & \checkmark & \checkmark & \checkmark & 58.2 & 55.4 & 36.7 & 49.3 \\
  (iv)  & adaptive   & \checkmark & \checkmark & \checkmark &            & \textbf{58.4} & \textbf{56.6} & \textbf{37.8} & \textbf{53.0} \\
  (v)   & adaptive  &            & \checkmark & \checkmark & \checkmark & 55.0 & 53.5 & 35.5 & 51.5 \\
  (vi)  & adaptive   & \checkmark &            & \checkmark & \checkmark & 58.3 & 56.7 & 37.7 & 52.6 \\
 (vii) & adaptive   & \checkmark & \checkmark &            & \checkmark & 58.4 & 56.3 & 37.6 & 52.4 \\
  (viii) & adaptive & \checkmark & \checkmark & \checkmark & \checkmark & \textbf{58.4} & \textbf{56.6} & \textbf{37.8} & \textbf{53.0} \\

\bottomrule
\end{tabular}
\caption{Ablation studies for reference-free setting. ``$\tau_b$'' and ``$\tau_c$'' represents the Kendall’s $\tau_b$ and $\tau_c$ correlation coefficient, respectively.}
\label{tab:sup_combined_ablation}
\end{table*}

\subsection{Human Performance on Nebula Dataset}
To evaluate human performance, we conducted a subject experiment on the Nebula dataset.
We opted for the Nebula dataset because of its reliability in ground-truth judgments; the ground-truth judgments are given by an average of four annotators per sample, a higher number than those of other standard datasets~\cite{composite, flickr}.
First, we randomly selected 330 samples from the test set.
Next, four subjects provided scores for these samples in a reference-free setting. 
We then calculated Kendall's $\tau$ for each subject’s judgments against the ground truth and computed the average across all subjects to measure human performance.
The results showed that the coefficients for the human performance were 63.3 and 63.0 for $\tau_b$ and $\tau_c$, respectively.

\section{Appendix C: Experimental Setup}
\paragraph{Baselines.}
In the reference-based setting, we employed BLEU~\cite{bleu}, ROUGE~\cite{rouge}, METEOR~\cite{meteor}, CIDEr~\cite{cider}, and SPICE~\cite{spice} as baseline metrics, as these are standard metrics for image captioning tasks. 
We also employed BERTScore~\cite{bertscore}, BARTScore~\cite{bartscore},  Ref-CLIPScore~\cite{clipscore}, SPARCS~\cite{smurf}, Ref-PAC-S~\cite{pac-s}, Ref-PAC-S++~\cite{sarto2024positive}, Ref-HICEScore~\cite{Zeng_2024hicescore}, Polos~\cite{polos}, \textsc{Deneb}~\cite{deneb}, FLEUR~\cite{lee2024fleur}, 
G-VEval~\cite{tong2024gveval}, and  HiFiScore~\cite{yao2024hifi} as baseline metrics, since they are representative reference-based metrics.

In the reference-free setting, we adopted CLIP-S~\cite{clipscore}, HICEScore~\cite{Zeng_2024hicescore}, PAC-S++~\cite{sarto2024positive}, BRIDGE~\cite{bridge}, BLIP2-Score~\cite{zeng2024meacap}, FLEUR~\cite{lee2024fleur},  HiFiScore~\cite{yao2024hifi}, and EXPERT~\cite{expert} as baselines  because they are representative reference-free metrics.
Note that CLAIR~\cite{chan2023clair}, FLEUR~\cite{lee2024fleur}, G-VEval~\cite{tong2024gveval}, HiFiScore~\cite{yao2024hifi}, and EXPERT~\cite{expert} are LLM-based metrics, while the others are LLM-free.

\begin{table*}[t]
\centering
\small 
\begin{tabular}{c c c c c c c c c}
\toprule
  \multirow{2}{*}{\textbf{Metric}} 
 & \multirow{2}{*}{\textbf{ARSM}}
 & \multicolumn{3}{c}{\textbf{Image Encoder}} 
 & \textbf{Text Encoder} 
 & \textbf{Composite}
 & \textbf{Flickr8K-Ex}
 & \textbf{Flickr8K-CF}\\
\cmidrule(lr){3-5} \cmidrule(lr){6-6} \cmidrule(lr){7-7} \cmidrule(lr){8-8} \cmidrule(lr){9-9}
 &  & CLIP & BLIP-2 & BEiT-3 & Stella
 & $\tau_c$ & $\tau_c$ & $\tau_b$\\
\midrule
\rowcolor{gray!10}
\multicolumn{9}{l}{\textbf{Reference-based}}\\
  (A)   & adaptive       & \checkmark &  &  & & 53.7 & 53.0 & 35.8  \\
   (B)& adaptive    & \checkmark & \checkmark & \checkmark & \checkmark & \textbf{60.4} & \textbf{58.6} & \textbf{38.6}  \\
\rowcolor{gray!10}
\multicolumn{9}{l}{\textbf{Reference-free}}\\

(C) & adaptive  & \checkmark &  &  &  & 51.6 & 51.1 & 33.8 \\
  (D)& adaptive   & \checkmark & \checkmark & \checkmark & \checkmark & \textbf{58.4} & \textbf{56.6} & \textbf{37.8} \\

 \bottomrule
\end{tabular}
\caption{Additional results of feature extractor ablation. We removed all encoders except the CLIP encoder to assess the impact of CLIP. ``$\tau_b$'' and ``$\tau_c$'' represents the Kendall’s $\tau_b$ and $\tau_c$ correlation coefficient, respectively.}
\label{tab:clip_ablation}
\end{table*}

\paragraph{Implementation details.}
We adopted the Adam optimizer ($\beta_1$ = 0.9, $\beta_2$ = 0.999) for training, with a learning rate of $10^{-5}$, a batch size of 16, and max number of epochs, 5. 
The number of CNN layers in HadamardNet was set to 8. The VSS submodule used 2 Transformer layers and 1 Q-Former layer.
We employed early stopping during training based on Kendall's $\tau_c$. Specifically, the value of $\tau_c$ was measured on the validation set at the end of each epoch. 
Training was stopped when the following condition was satisfied: $\tau_c^{(t)} \leq \tau_c^{(t-1)}$, where $\tau_c^{(t)}$ denotes the value of $\tau_c$ at the end of epoch $t$ on the validation set. After training, the model with the highest $\tau_c$ value on the validation set was selected, and its performance was evaluated on the test set.

Following previous works \cite{clipscore, pac-s, polos,deneb, lee2024fleur, yao2024hifi}, all experiments were reported based on a single run.
Our model had approximately 119 million trainable parameters.
We trained our model on a GeForce RTX 3090 with 24GB of memory and an Intel Core i9 12900K with 64GB of memory.
The training phase was completed in approximately 4.5 hours, and the inference time was approximately 8.2 ms/sample.

\section{Appendix D: Additional Experiments}

\subsection{Additional Comparison with Baselines}

Table~\ref{tab:additional_baseline} shows a quantitative comparison with additional baselines, including BERTScore \cite{bertscore}, BARTScore \cite{bartscore}, SPARCS \cite{smurf}, and BRIDGE \cite{bridge}, which were omitted from the main body due to space constraints.
These results demonstrated that Pearl consistently outperformed these baselines on the Composite, Flickr8K-Expert, Flickr8K-CF, and Nebula datasets, demonstrating its superior effectiveness.

\subsection{Additional Ablation Studies}
\paragraph {ARSM ablation in reference-free settings.}
\Cref{tab:sup_combined_ablation} shows the ARSM ablation and the feature extractor ablation in reference-free setting. Metric (i) yielded correlation coefficients of 30.7, 30.4, 16.7, and 42.3 for Composite, Flickr8K-Expert, Flickr8K-CF, and Nebula, respectively . 
These values decreased by 27.7, 26.2, 21.1, and 10.7  points compared with those of Metric (viii).

Moreover, Metric (iii)
yielded correlation coefficients 
of 
58.2, 55.4, 36.7 and 49.3
for Composite, Flickr8K-Expert, Flickr8K-CF, and Nebula, respectively. 
These values decreased by 
0.2, 1.2, 0.9, and 3.7 points compared with those of Metric (viii). 
These results indicate that ARSM was effective at extracting significant features for automatic evaluation and significantly contributed to performance in reference-free settings.

\paragraph{Feature extractor ablation in reference-free settings.}
We investigated the contribution of image and text encoders by removing CLIP, BLIP-2, BEiT-3, and Stella in the reference-free settings.
\Cref{tab:sup_combined_ablation} demonstrates the correlation coefficients in the reference-free setting. Using Metric (v) were 55.0, 53.5, 35.5 and 51.5 on Composite, Flickr8K-Expert, Flickr8K-CF, and Nebula, respectively.
These values marked decreases of 3.4, 3.1, 2.3 and 1.5 points compared with those of Metric (viii).
These results indicate that CLIP significantly contributed to the performance of Pearl in reference-free settings.

\begin{table}[t]
\centering
\setlength{\tabcolsep}{1mm} 
\begin{tabular}{llcccc}
\toprule

\multirow{2}{*}{\textbf{Metrics}} &  \multirow{2}{*}{\textbf{Training} \textbf{Dataset}}  & \textbf{Com} & \textbf{Ex} & \textbf{CF}  \\  
\cmidrule(lr){3-3} \cmidrule(lr){4-4} \cmidrule(lr){5-5} 
& & $\tau_c$ & $\tau_c$ & $\tau_b$  \\ \midrule
\rowcolor{gray!10}
\multicolumn{5}{l}{\textbf{Reference-based}}\\
(A)  & Nebula   & 59.2  & 57.7 & 37.9  \\
(B)  & Ref-based Spica  & 59.9 &  56.2 & 38.3 \\
(C)  & Spica  & \textbf{60.4}  & \textbf{58.6} & \textbf{38.6}  \\
\midrule

\rowcolor{gray!10}
\multicolumn{5}{l}{\textbf{Reference-free}}\\
(D)  & Nebula   &57.4  & 51.7 & 37.1  \\
(E)  & Ref-free Spica  & 57.5 & 56.2 & 37.7 \\
(F)  & Spica   & \textbf{58.4}  & \textbf{56.6} & \textbf{37.8}  \\

\bottomrule
\end{tabular}
\caption{Results of Spica ablation and single-model ablation. ``Com'', ``EX'' and ``CF'' represent Composite,
Flickr8K-Ex and Flickr8K-CF, respectively.}
\label{tab:dataset_ablation_sup}
\end{table}

As aforementioned in the \textit{Experiments} section, these results do not indicate that CLIP is the sole contributor to Pearl’s performance.
To further investigate this, we conducted an additional ablation study.
Table~\ref{tab:clip_ablation} shows the results of the ablation study in which all encoders except the CLIP encoder were removed to assess the impact of CLIP.
Compared to the original Pearl (B) and (D), the CLIP-only variants (A) and  (C) exhibit a significant performance drop.
Results from Tables~\ref{tab:sup_combined_ablation} and~\ref{tab:dataset_ablation_sup} indicate that the combination of CLIP, BLIP-2, BEiT-3, and Stella encoders contributed substantially to the overall performance.

\paragraph{Spica ablation}
We evaluated the effectiveness of the Spica dataset by training an existing supervised metric on Spica in both reference-based and reference-free settings.
Table~\ref{tab:dataset_ablation_sup} presents the results of the Spica ablation.
As shown in this table, Pearl trained on Spica (Metrics (C) and (F)) outperformed the model trained on Nebula (Metrics (A) and (D)) in both settings.
These results demonstrate the contribution of the Spica dataset to overall performance.

\paragraph{Single-model ablation.}
Pearl adopts a single-model strategy that jointly addresses both settings because this design allows all samples -- whether reference-based or reference-free -- to contribute to training (See the \textit{Methodology} section.)
To evaluate the effectiveness of this strategy, we split Spica into two subsets based on the presence or absence of reference captions and trained Pearl on each subset.
Specifically, we created a ``Ref-based Spica'' subset containing only samples with references, and a ``Ref-free Spica'' subset containing only samples without references.
Table~\ref{tab:dataset_ablation} presents the results of this single-model ablation on Composite, Flickr8K-EX, and Flickr8K-CF.
Compared to the original Pearl (C), the variant trained on Ref-based Spica (B) shows a performance drop across all benchmarks, with the largest decrease observed on Flickr8K-Expert (-2.0 points).
Similarly, the variant trained on Ref-free Spica (E) also underperformed compared to the original model (F).
These results indicate that the single-model strategy contributed to overall performance.

\section{Appendix E: Error Analysis}
\label{sup:error_analysis}

\begin{table}[t]
\centering
\begin{tabular}{lc} \toprule
Description           & \#Error \\ \midrule
Descriptiveness Error & 42      \\
Relevance Error       & 40      \\
Annotation Errors     & 9       \\
Fluency Error         & 5       \\
Others                & 4       \\ \midrule
Total                & 100    \\ \bottomrule
\end{tabular}
\caption{Categorization of failure modes. We analyzed the 100 samples with the greatest absolute differences between $\hat{y}$ and $y$.}
\label{tab:error}
\end{table}

The effectiveness of our metric was validated on standard image captioning benchmarks; however, it does come with limitations. To examine these limitations, we analyzed the 100 samples with the greatest absolute differences between $\hat{y}$ and $y$.
\Cref{tab:error} summarizes the failure modes into five primary categories:
\begin{itemize}
    \item \underline{Descriptiveness error}: This category includes samples where our metric assigned incorrect scores to captions that lacked descriptiveness.
    \item \underline{Relevance error}: This category includes samples where the metric assigned inappropriate scores to candidates with incorrect details.
    \item \underline{Annotation errors}: This category includes samples where the human judgments were inappropriate.
    \item \underline{Fluency error}: This category includes samples where the metric assigned inappropriate scores to captions containing grammatical errors.
    \item \underline{Others}: This category includes miscellaneous errors that do not fall into the above categories.
\end{itemize}

\end{document}